\documentclass{article}
\pdfoutput=1
\usepackage[T1]{fontenc}
\usepackage[utf8]{inputenc}
\usepackage{amsmath}
\usepackage{amssymb}
\usepackage{amsfonts}
\DeclareUnicodeCharacter{2248}{\ensuremath{\approx}} 
\DeclareUnicodeCharacter{2192}{\ensuremath{\rightarrow}} 
\DeclareUnicodeCharacter{2013}{--} 
\DeclareUnicodeCharacter{2014}{---} 
\DeclareUnicodeCharacter{2019}{'} 
\DeclareUnicodeCharacter{2018}{`} 
\DeclareUnicodeCharacter{201C}{``} 
\DeclareUnicodeCharacter{201D}{''} 
\pdfoutput=1
\usepackage{spconf,amsmath,graphicx,hyperref}
\usepackage{booktabs}
\usepackage{multirow}

\title{Confidence-Gated Training for Efficient Early-Exit Neural Networks}
\name{Saad Mokssit$^{\star}$ \qquad Ouassim Karrakchou$^{\star}$ \qquad Alejandro Mousist $^{\dagger}$ \qquad Mounir Ghogho$^{\ast \ddagger}$\thanks{This research has received funding from the European Union’s Horizon research and innovation program under grant agreement No~101070374.}}
\address{$^{\star}$International University of Rabat,	TICLab, Morocco\\
$^{\dagger}$ Thales Alenia Space, Tres Cantos, Spain  \\
$^{\ast}$ Mohammed 6 Polytechnic University, College of Computing, Rabat, Morocco \\
$^{\ddagger}$  University of Leeds, Faculty of Engineering, United Kingdom        }
\begin{document}
%
\maketitle
\begin{abstract}
Early-exit neural networks reduce inference cost by enabling confident predictions at intermediate layers. However, joint training often leads to gradient interference, with deeper classifiers dominating optimization. We propose Confidence-Gated Training (CGT), a paradigm that conditionally propagates gradients from deeper exits only when preceding exits fail. This encourages shallow classifiers to act as primary decision points while reserving deeper layers for harder inputs. By aligning training with the inference-time policy, CGT mitigates overthinking, improves early-exit accuracy, and preserves efficiency. Experiments on the Indian Pines and Fashion-MNIST benchmarks show that CGT lowers average inference cost while improving overall accuracy, offering a practical solution for deploying deep models in resource-constrained environments.
\end{abstract}
\section{Introduction}
\label{sec:intro}

Deep neural networks (DNNs) deliver state-of-the-art results in vision, language, and video \cite{wang2021comparative,otter2020survey,oprea2020review}, but their accuracy often comes with heavy compute and energy costs \cite{thompson2020computational}, limiting deployment on mobile, satellite, and edge platforms \cite{chen2019deep}. Early-exit networks \cite{sun2021early,rahmath2024early,campbell2022robust} address this by attaching internal classifiers to intermediate layers and terminating inference at the first exit whose confidence exceeds a threshold. This design routes easy inputs to shallow, low-cost predictions while allowing harder ones to proceed deeper, thereby allocating computational resources adaptively on a per-sample basis.


Training such models is inherently multi-objective: each exit $e$ defines its own loss $\mathcal{L}_e$ for achieving high accuracy at depth $e$. The prevailing approach collapses this vector objective $(\mathcal{L}_1, \mathcal{L}_2,\dots,\mathcal{L}_E)$ into a fixed, exit-weighted scalarization $\mathcal{L} = \sum_{e} \lambda_e \mathcal{L}_e$ with static hyperparameters $\lambda_e$ \cite{kaya2019shallow,matsubara2022split}. The resulting gradient $g=\sum_{e}\lambda_e\nabla\mathcal{L}_e$ may exhibit gradient conflict when exit-level gradients point in different directions. Because the $\lambda_e$’s are fixed and input-agnostic, the same trade-off is imposed on {\em all} samples during training, regardless of their difficulty. In practice, losses from deeper exits tend to dominate, leaving shallow exits under-optimized and inducing the ‘overthinking’ effect on easy samples \cite{kaya2019shallow}. This occurs when inputs that could be classified at shallow exits still backpropagate gradients from deeper heads, causing unnecessary updates that perturb early-layer representations already sufficient for these easy examples.  Classifier-only training \cite{sun2021early}, cascaded training \cite{luo2024cess}, and distillation/boosting variants \cite{phuong2019distillation,banijamali2023pyramid} all treat samples equally and therefore suffer from the same issues described above.

Our perspective connects naturally to methods that modulate learning at the sample level. Adaptive Focal Loss down-weights easy, high-confidence examples and emphasizes harder ones \cite{talamantes2024transunet}; curriculum and self-paced learning reweight or schedule examples by difficulty \cite{gu2023accelerating}; and sample reweighting methods adapt weights based on past errors \cite{fu2024learning}. However, these approaches are not exit-aware: they adjust example importance globally but do not couple the weight of a loss term to the exit at which an example would terminate in a multi-exit network. Closest to our work is the CLassyNet framework \cite{ayyat2023classynet}, but it operates on a class basis rather than a sample basis: shallow heads are trained to recognize only predefined class groups, so exit decisions depend primarily on class membership rather than per-sample difficulty.

We introduce \textbf{Confidence Gated Training (CGT)}, which replaces fixed exit weights $\lambda_e$ with sample-dependent weights $\lambda_e^{(i)}$ so as to mimic the adaptive processing at inference. We present two instantiations of this approach: \textbf{HardCGT} and its soft variant \textbf{SoftCGT}.
In HardCGT, a binary eligibility mask $\lambda_e^{(i)}\!\in\!\{0,1\}$ allows deeper exits to receive gradients only if earlier exits fail for sample $i$ (due to either low confidence or incorrect classification). SoftCGT generalizes this with a continuous residual gate $\lambda_e^{(i)}\!\in\![0,1]$ that scales deeper-exit gradients in proportion to the residual uncertainty left by earlier exits. Viewed through a multi-objective lens, Hard and Soft CGT perform a dynamic, sample-conditioned scalarization that better tracks the desired Pareto trade-off: easy inputs shape shallow classifiers; hard inputs sustain learning at deeper exits. This turns early exiting from a heuristic applied only at inference into an active, sample-adaptive training principle that reduces gradient conflict and balances accuracy–efficiency across exits.

\section{Overthinking in early exit DNNs}
\begin{figure*}[htb]
\begin{minipage}[b]{1.0\linewidth}
  \centering
  \centerline{\includegraphics[width=12cm]{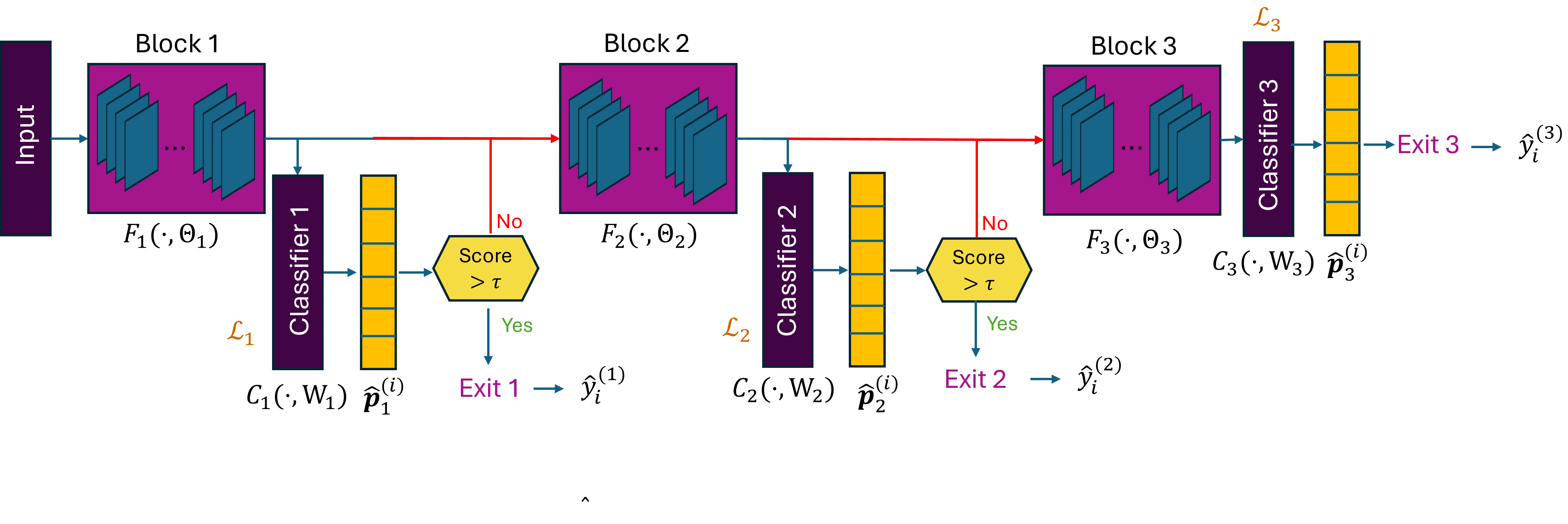}}
\end{minipage}
\vspace{-0.4cm}
\caption{Early exit DNN. The input flows through successive backbone blocks. At each block a head outputs a prediction and confidence. If the confidence exceeds the threshold, inference terminates, otherwise the features continue to the next block.}
    \label{fig:early exit network}
\end{figure*}
Consider supervised multi-class classification with a dataset $\mathbb{D}=\{(\boldsymbol{x}_i,y_i)\}_{i=1}^N$, where $\boldsymbol{x}_i$ is the $i$th input and $y_i\in\{1,\ldots,C\}$. An early-exit network (see fig.\ref{fig:early exit network}) augments a NN-based classifier with intermediate classification modules that allow inference to terminate early when confidence is high.

We model the backbone of the DNN as a composition of $E$ sequential blocks,
each parameterized by a set of parameters $\boldsymbol{\theta}_e$, for $e=1,\cdots,E$. For each exit $e$, we attach to the backbone a classification module, parameterized by the set of parameters $\boldsymbol{W}_e$, which produces a vector of class probabilities $\boldsymbol{p}_e^{(i)} = \big[p_{e,1}^{(i)}, \ldots, p_{e,C}^{(i)}\big]^T$,
where $p_{e,c}^{(i)}$ denotes the probability that input $\boldsymbol{x}_i$ belongs to class $c$ at exit $e$.


The training objective is to jointly learn the backbone and the classification modules so that, at inference, easy samples can be processed by early classifiers while harder samples are passed to deeper ones. However, training early-exit DNNs is challenging. Since branch $e$ shares parameters with all preceding branches, the gradient with respect to any shared block 
aggregates contributions from all subsequent exits. This naturally induces a multi-objective optimization problem, as gradients from different exit losses $\mathcal{L}_e$ may conflict. A common strategy to address this challenge is scalarization, which minimizes a weighted sum of the per-exit losses, i.e., we minimize
\begin{equation}
    \mathcal{L}(\boldsymbol{\theta},\boldsymbol{W})=\sum_{e=1}^{E}\lambda_e\,\mathcal{L}_e(\boldsymbol{\theta}_{1:e},\boldsymbol{W}_e)=\frac{1}{N}\sum_{e=1}^{E}\lambda_e \sum_{i=1}^{N}\ell\!\left(\hat{\boldsymbol{p}}^{(i)}_e,\,y_i\right),
    \label{joint loss}
\end{equation}
where $\boldsymbol{\theta}_{1:e}=\{\boldsymbol{\theta}_1, \cdots,\boldsymbol{\theta}_e\}$, $\ell(\cdot,\cdot)$ denotes the cross-entropy loss, and the $\lambda_l$ are positive-valued hyperparameters that can be tuned via cross-validation. Importantly, validation may consider not only accuracy but also computational cost, as the rationale behind early exiting is to reduce inference time and, consequently, energy consumption.
 

The fixed scalarization (i.e., fixed values for the $\{\lambda_e\}$) enforces the same trade-off across all samples, regardless of which exit will process them at inference. As a result, deeper, higher-capacity heads may dominate the shared parameter updates. This makes early branches difficult to optimize, as they receive weaker or even conflicting signals through shared parameters, leading to under-trained shallow exits. This phenomenon can be seen as a form of \emph{training overthinking}, where early representations are shaped primarily as intermediates for deeper exits rather than as reliable predictors for easy samples. Optimizing the fixed hyperparameters $\lambda_e$ alleviates this issue to some extent, but the uniform treatment of all samples during training maintains a mismatch with the inference process. The proposed method mitigates this limitation by making the scalarization weights input-dependent, enabling a more adaptive balance between early and late exits and aligning the training process more closely with inference.

\section{Confidence Gated Training}
The overthinking problem motivates training schemes that account for the multi-objective nature of early exiting and modulate learning signals to mitigate gradient conflicts on shared blocks, ideally in a sample-aware manner that reflects the intended exit behavior at inference. To this end, we propose Confidence-Gated Training (CGT), which minimizes the following joint loss:
\begin{equation}
    \mathcal{L}_{\text{CGT}} = \frac{1}{N} \sum_{i=1}^N \sum_{e =1}^E \lambda_e^{(i)}  \ell\!\left(\hat{\boldsymbol{p}}_e^{(i)}, y_i\right),
\end{equation}
Here, the weights are input-dependent, unlike the fixed formulation in Eq.~(\ref{joint loss}). CGT adjusts the effective per-exit weights on a per-sample basis, increasing the influence of exits that misclassify a sample while reducing or suppressing gradients from deeper exits when earlier ones are both confident and correct. This sample-conditioned gradient control prevents unnecessary transformations of already discriminative features and aligns training with the early-exit inference policy. As a result, shallow exits are explicitly optimized to handle easy inputs, while deeper layers specialize in difficult or ambiguous cases, thereby mitigating overthinking and improving computational efficiency. Gradient conflict still persists, as it is inherent to early-exit DNNs, but its impact is reduced compared to the conventional fixed-weight scalarization approach. In the following, we present two instantiations of the CGT approach, which consist of different designs of the input-dependent scalarization weights. 

\subsection{Hard CGT}
The central idea is to allow gradients from a deeper exit $e$ to propagate only when all preceding exits fail to produce a confident and correct prediction. This ensures that earlier exits serve as the primary decision points for simpler inputs, while deeper exits are reserved for harder cases. We implement this idea as follows.

For input $i$, we define a success indicator at exit $e$ as:
\begin{equation}
    \delta_e^{(i)} =
    \begin{cases}
        1 & \text{if } \hat{y}_i^{(e)} = y_i \ \text{and} \ s_e^{(i)} \geq \tau, \\
        0 & \text{otherwise,}
    \end{cases}
\end{equation}
where $\hat{y}_i^{(e)}$ is the label estimate at exit $e$, i.e. \begin{equation}
    \hat{y}_i^{(e)} = \arg\max_{c \in \{1,\ldots,C\}} \hat{p}_{e,c}^{(i)},
\end{equation}
$s_e^{(i)} = \max_c \hat{p}_{e,c}^{(i)}$ is the associated confidence score, and $\tau \in (0,1)$ is a predefined confidence threshold.

The eligibility of gradients is then determined by
\begin{equation}
    \lambda_e^{(i)} = 1,\quad\lambda_e^{(i)} = \prod_{e'<e}\big(1 - \delta_{e'}^{(i)}\big) ~ {\rm for ~ }e>1
    \label{binary mask}
\end{equation}
i.e. $\lambda_e^{(i)}$ equals $1$ only if all earlier exits $e'<e$ have failed.  

\subsection{Soft CGT}
While HardCGT effectively prioritizes shallow exits by blocking gradients from deeper classifiers whenever an earlier exit succeeds, its binary gating mechanism can cause limitations. In particular, deeper exits may receive progressively fewer training samples as shallow classifiers become increasingly confident, leading to a sample starvation problem. This restricts deeper exits from refining their representations for harder inputs and creates abrupt changes in the optimization signal.

To address this, we extend HardCGT with Residual Gating, a continuous and confidence-aware mechanism that modulates gradient flow rather than blocking it outright. The key idea is that the amount of gradient passed to deeper exits should be proportional to the residual uncertainty left unresolved by earlier exits. High-confidence correct predictions at shallow exits strongly attenuate gradients flowing deeper, while uncertain or incorrect predictions permit stronger gradient flow, enabling deeper layers to specialize on more difficult cases.

We define a residual gating coefficient for exit $e$ as:
\begin{equation}
r_e^{(i)} = 1 - \sigma(s_e^{(i)} - \tau),
\end{equation}
where $\sigma(\cdot)$ is the sigmoid function. When $s_e^{(i)} \ll \tau$, the gate $r_e^{(i)} \approx 1$, allowing strong gradients to flow; when $s_e^{(i)} \gg \tau$, the gate $r_e^{(i)} \approx 0$, attenuating gradients from deeper exits.

The cumulative residual gating factor for exit $e$ is then defined as:
\begin{equation}
\lambda_e^{(i)} = 1,\quad \lambda_e^{(i)} = \prod_{e'<e} r_{e'}^{(i)} ~ {\rm for ~ }e>1.
\end{equation}
Unlike the binary eligibility mask in HardCGT (Eq.~\ref{binary mask}), the cumulative residual gating factor provides a smooth, sample-dependent mechanism for controlling gradient propagation. This formulation preserves the principle of CGT by prioritizing shallow exits, while ensuring that deeper exits continue to receive adequate training signals. By modulating gradient flow softly rather than suppressing it entirely, SoftCGT promotes more stable optimization, balanced exit training, and improved robustness across input difficulty levels.


\section{Experiments}

We evaluate on two complementary benchmarks and use architectures matched to each domain. Indian Pines \cite{vishwanath2024hyperspectral} is an AVIRIS hyperspectral scene of size $145 \times 145$ with 200 effective bands and 16 classes. We perform pixel-wise classification by masking labeled pixels, vectorizing 200-D spectra, min–max normalizing features, and applying a stratified 70/30 train–test split. Fashion-MNIST \cite{xiao2017fashion} comprises 70,000 grayscale $28 \times 28$ images from 10 categories (60k/10k split) with per-channel normalization (mean 0.286, std 0.353) and light augmentation (random crop + horizontal flip). We compare Hard/Soft CGT with three baselines—BranchyNet\cite{teerapittayanon2016branchynet}, Cascade Optimization (CO) \cite{luo2024cess}, and ClassyNet \cite{ayyat2023classynet}. For each baseline, exit weights are tuned by grid search under the constraint $\lambda_1> \lambda_2> \lambda_3 $ to prevent degenerate solutions that concentrate most of the weight to the last exit. All methods share the same exit placement per dataset. For Indian Pines, the backbone is a 5-layer MLP (width 256) with BatchNorm, ReLU, and dropout $p = 0.2$; exits are attached at layers $\{1,3,5\}$. For Fashion-MNIST, the backbone is a lightweight CNN with three blocks with channel widths $\{32,64,128\}$, each block being $3 \times 3$ Conv $\rightarrow$ BatchNorm $\rightarrow$ ReLU $\rightarrow 2\times 2$ MaxPool, and a global average pool feeding three exits. Inference adopts early exiting with  confidence threshold $\tau =0.9$.

\section{Results and Discussion}

\begin{table*}[]
\begin{tabular}{@{}ccclllllll@{}}
\toprule
\multicolumn{1}{l}{Dataset}    & \multicolumn{1}{l}{Task}              & \multicolumn{1}{l}{Backbone} & Model   & F1            & Precision     & Recall        & \#Exit1         & \#Exit2         & \#Exit3         \\ \midrule
\multirow{5}{*}{Indian Pines}  & \multirow{5}{*}{Segmentation}   & \multirow{5}{*}{MLP}         & BranchyNet \cite{teerapittayanon2016branchynet}     & 88\%          & 89\%          & 88\%          & 35\%            & 23.3\%          & 41.7\%          \\
                               &                                       &                              & CO\cite{luo2024cess}      & 71\%          & 87\%          & 69\%          & 57\%            & 0\%             & 43\%            \\
                               &                                       &                              & ClassyNet\cite{ayyat2023classynet}     & 94.5\%        & 94.5\%        & 94.5\%         & 0.1\%             & 45.6\%            & 54.3\%           \\
                               &                                       &                              & HardCGT & 92\%          & 93\%          & 92\%          & \textbf{64\%}   & 15.5\%          & 20.5\%          \\
                               &                                       &                              & SoftCGT & \textbf{95\%} & \textbf{96\%} & \textbf{95\%} & 60\%            & 21.3\% & \textbf{18.7\%} \\ \midrule
\multirow{5}{*}{Fashion Mnist} & \multirow{5}{*}{Classification} & \multirow{5}{*}{CNN}         & BranchyNet \cite{teerapittayanon2016branchynet}     & 89\%          & 89\%          & 89\%          & 6.8\%           & 32.8\%          & 60.4\%          \\
                               &                                       &                              & CO\cite{luo2024cess}      & 86\%          & 86\%          & 86\%          & 4.8\%           & 19.3\%          & 75.9\%          \\
                               &                                       &                              & ClassyNet\cite{ayyat2023classynet}     & \textbf{94\%}          & \textbf{94\%}          & \textbf{94\%}          & 0.5\%             & 57.8\%           & \textbf{41.7\%}          \\
                               &                                       &                              & HardCGT & 90\%          & 90\%          & 90\%          & 7.5\%           & 30.6\%          & 61.9\%          \\
                               &                                       &                              & SoftCGT & 91.5\%        & 91.5\%        & 91.5\%        & \textbf{10.8\%} & 40.3\% & 48.9\% \\ \bottomrule
\end{tabular}
\caption{Accuracy and routing for early-exit training strategies on Indian Pines and Fashion-MNIST. \#Exit k gives the percentage of test samples that terminate at exit k (lower \% at deeper exits implies earlier, cheaper inference).}
\label{models eval}
\end{table*}






\begin{figure}[htb]

\begin{minipage}[b]{1.0\linewidth}
  \centering
  \centerline{\includegraphics[width=7.5cm]{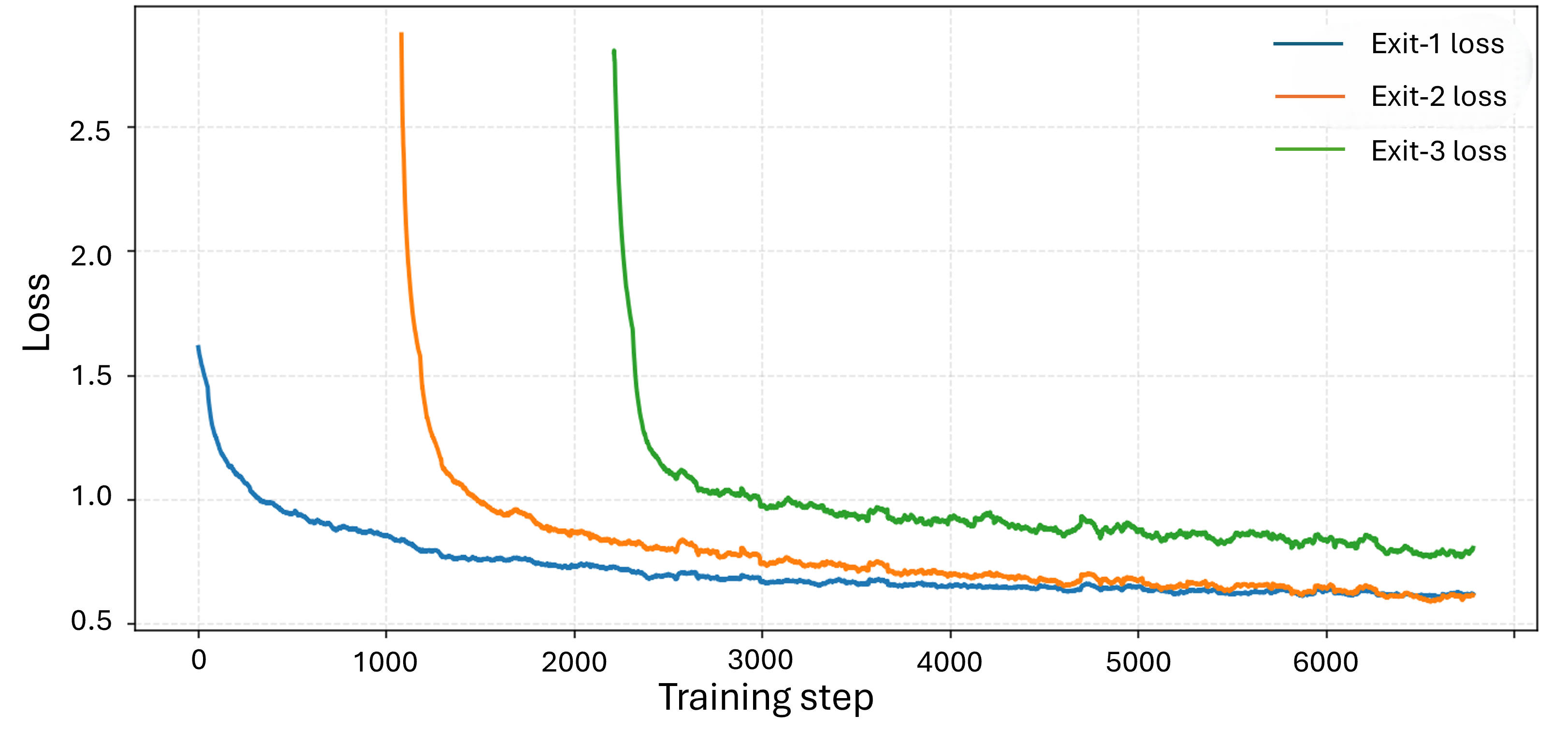}}
\end{minipage}

\caption{Per exit training loss under HardCGT}
    \label{fig:train_ef}
\end{figure}

\begin{figure}[htb]

\begin{minipage}[b]{1.0\linewidth}
  \centering
  \centerline{\includegraphics[width=7.5cm]{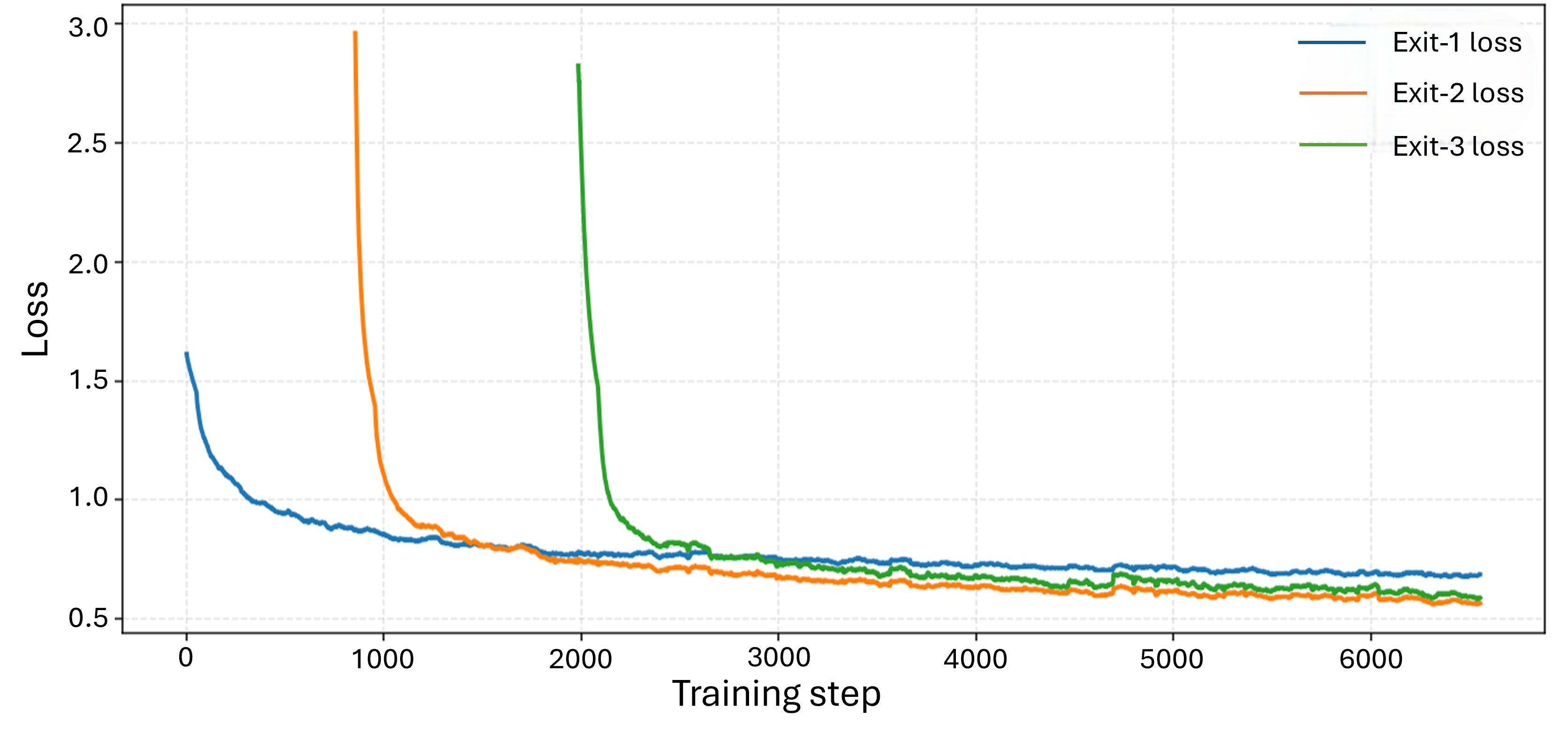}}
\end{minipage}

\caption{Per exit training loss under SoftCGT.}
    \label{fig:train_efrg}
\end{figure}

Table \ref{models eval} compares five training regimes on Indian Pines and Fashion-MNIST. On Indian Pines, SoftCGT achieves the best accuracy (F1/Prec/Rec = 95/96/95\%) with efficient routing (60/21.3/18.7\% at Exits 1–3); HardCGT is close (92/93/92\%) and routes earlier (64/15.5/20.5\%). ClassyNet is competitive (94.5\%) but defers computation to deeper layers (0.1/45.6/54.3\%), reflecting its design where shallow heads specialize on fixed class groups, so exit decisions depend on class membership rather than per-sample difficulty. BranchyNet trails (~88–89\%) with more uniform traffic (35/23.3/41.7\%), and CO is lowest (71/87/69\%) with an under-used mid exit (57/0/43\%). On Fashion-MNIST, ClassyNet leads (94\%) but again relies on deeper computation (0.5/57.8/41.7\%), whereas SoftCGT (91.5\%) and HardCGT (90\%) strike a better accuracy–efficiency balance by shifting more samples earlier (SoftCGT 10.8/40.3/48.9\%, HardCGT (7.5/30.6/61.9\%). BranchyNet spreads traffic, while CO pushes most samples to the final exit.

The per-exit training curves in Figures \ref{fig:train_ef} and \ref{fig:train_efrg} explain SoftCGT’s gains. When a new head appears, both settings show the expected ‘“cold-start spike → rapid decay.’” Under HardCGT, Exit-1/2 converge near 0.62–0.65, but Exit-3 stalls around ≈0.80—consistent with sample starvation once shallow exits become confident. SoftCGT mitigates this by supplying depth-proportional gradients: Exit-3 drops promptly after activation and converges to ≈0.60–0.62, closing the gap to Exit-2 and even undercutting Exit-1. SoftCGT’s curves are also smoother late in training, indicating more stable optimization. The trade-off is a modestly higher Exit-1 loss (≈0.68–0.70 vs. ≈0.62 with HardCGT), reflecting that some difficult examples are handled at deeper exits. Overall, the curves support the residual-gating hypothesis: preserved shallow discriminability, no starvation at depth, and more balanced learning across exits.

\section{Conclusion}
In this work, we introduced Confidence-Gated Training (CGT), a paradigm for multi-exit networks that explicitly aligns optimization with the inference-time objective of early exiting. HardCGT prioritizes shallow exits by suppressing gradient flow from deeper layers when earlier classifiers are confident and correct, while SoftCGT adaptively modulates this suppression to ensure that deeper exits remain sufficiently trained. Experiments on the Indian Pines and Fashion-MNIST benchmarks demonstrate that CGT reduces average inference cost while improving accuracy compared to existing strategies. By making early exiting an explicit training objective, CGT establishes a principled framework for efficient and reliable dynamic inference, with future work aimed at exploring learned gating mechanisms, integration with uncertainty calibration, and extensions to tasks beyond classification.

\newpage


\bibliographystyle{IEEEbib}
\bibliography{refs}

\end{document}